\documentclass[a4paper]{article}

\usepackage{authblk}
\usepackage[utf8]{inputenc}
\usepackage[T1]{fontenc}
\usepackage{pslatex}
\usepackage[english]{babel}
\usepackage{fancyhdr}
\usepackage{graphicx}
\usepackage{changes}
\usepackage{float}
\usepackage{enumitem}
\usepackage{algpseudocode,algorithm,algorithmicx}
\usepackage[justification=centering]{caption}
\usepackage[justification=centering]{subcaption}
\usepackage{amsmath}
\usepackage{booktabs}
\usepackage{microtype}

\newcommand*\Let[2]{\State #1 $\gets$ #2}
\algrenewcommand\algorithmicrequire{\textbf{Precondition:}}
\algrenewcommand\algorithmicensure{\textbf{Postcondition:}}

\algblockdefx[Foreach]{Foreach}{EndForeach}[1]{\textbf{for each} #1 \textbf{do}}{\textbf{end for}}

\providecommand{\keywords}[1]{\textbf{\textit{Keywords }} #1}

\begin{document}

\title{NASCTY: Neuroevolution to Attack Side-channel
Leakages Yielding Convolutional Neural Networks}

\author[1]{Fiske Schijlen}
\author[1]{Lichao Wu}
\author[2]{Luca Mariot}

\affil[1]{{\normalsize Delft University of Technology, Mekelweg 5, 2628 CD, Delft, The Netherlands} \\

    {\small \texttt{fiske.schijlen@hotmail.com, l.wu-4@tudelft.nl}}}

\affil[2]{{\normalsize Semantics, Cybersecurity \& Services (SCS), University of Twente, Drienerlolaan 5, 7522NB Enschede, The Netherlands} \\

{\small \texttt{l.mariot@utwente.nl}}}
	
\maketitle

\begin{abstract}
Side-channel analysis (SCA) can obtain information related to the secret key by exploiting leakages produced by the device. Researchers recently found that neural networks (NNs) can execute a powerful profiling SCA, even on targets protected with countermeasures. This paper explores the effectiveness of \textit{Neuroevolution to Attack Side-channel Traces Yielding Convolutional Neural Networks} (NASCTY-CNNs), a novel genetic algorithm approach that applies genetic operators on architectures' hyperparameters to produce CNNs for side-channel analysis automatically. The results indicate that we can achieve performance close to state-of-the-art approaches on desynchronized leakages with mask protection, demonstrating that similar neuroevolution methods provide a solid venue for further research. Finally, the commonalities among the constructed NNs provide information on how NASCTY builds effective architectures and deals with the applied countermeasures.
\end{abstract}


\section{Introduction}
\label{sec:intro}
Cryptographic algorithms are a ubiquitous part of modern life since they allow us to preserve the confidentiality and integrity of sensitive data. However, the implementation of such algorithms (even if they are mathematically secure) can sometimes leak information about security assets, for instance, through power~\cite{Kocher_SCA} or electromagnetic radiation~\cite{10.1007/3-540-45418-7_17,dpa_book}. An attacker can attempt a \textit{side-channel analysis} (SCA) on leakages to exploit that leakage and retrieve the secret key or its parts. 

Assuming that the attacker has an identical copy of the target device, \textit{profiling} SCA becomes one of the most potent attack methods. Such an attack leverages traces generated on the copy to construct a model that profiles leakage patterns corresponding to the key-related intermediate data or the key itself. The profiling model can then recover the secret key from traces generated by the target device. Nowadays, neural networks (NNs) have become one of the most popular profiling model options thanks to their strong attack capability, even if the leakage traces are protected with countermeasures~\cite{ascad-paper,noise-cnss-sca,ae-for-sca-traces,efficient-cnn-arch-method,revisiting-efficient-cnn-archs}. This type of attack is commonly referred to as deep learning-based SCA (DL-SCA)~\cite{picek-sok}.

In practice, one of the biggest obstacles to applying DL-SCA is the design of the NN architecture and the optimization of its hyperparameters. The \textit{architecture} of an NN refers to its inner components, such as the neurons and the connections in between. In side-channel analysis research, an NN's architecture is often decided empirically, resulting in different architectures even on the same dataset~\cite{ascad-paper,efficient-cnn-arch-method,revisiting-efficient-cnn-archs}. Indeed, it is challenging to find an optimal architecture given an enormous number of hyperparameter combinations. Even worse, the selected architecture may not be transferrable when attacking different datasets or implementations. Therefore, it would be helpful to have a sophisticated and automated approach to build an architecture for an SCA on any given dataset. While there are other approaches to the automated design for neural networks for SCA, they also come with specific issues. For instance, Rijsdijk et al. used reinforcement learning that produced top-performing neural networks, but the authors still needed to start with a general description of architectures to be designed~\cite{Rijsdijk_Wu_Perin_Picek_2021}.
Additionally, a reinforcement learning approach is computationally expensive and requires a cluster of GPUs and days of tuning time. On the other hand, Wu et al. used Bayesian optimization to find neural network architectures for SCA~\cite{lichao-auto-hyperparam-tuning}. This approach is much faster than reinforcement learning while providing similar results (in terms of attack performance). Still, the authors needed to select the surrogate model and acquisition function for Bayesian optimization, which can again make the hyperparameter tuning significantly harder. Besides, both aforementioned methods rely on the experience obtained from iterations, and the question ``\textit{Is the selected model global optimal?''} is tricky to answer.

In this paper, we propose a \textit{Genetic Algorithm} (GA) as an alternative to the above-mentioned methods for the hyperparameter tuning task in the context of DL-SCA. In general, GAs are quite a versatile metaheuristic for hyperparameter tuning, which optimize a population of candidate hyperparameter vectors (or individuals, in the GA terminology). A GA mimics natural evolution for these vectors by applying genetic operators such as recombination and mutation, pruning the population with a selection method, and evaluating them against a fitness function. This process is iterated over multiple generations, after which the last generation's best-performing hyperparameter vector is taken as a solution. In principle, a GA allows for obtaining robust models for leakages acquired from different cryptographic implementations.

The main contributions of this work are:
\begin{enumerate}
\item We provide a methodology based on genetic algorithms for tuning the hyperparameters of neural networks for profiling side-channel analysis. Our approach is automated, extensible, and capable of producing various neural networks.
\item We analyze the components of well-performing architectures constructed with our method, giving insights into the effectiveness of CNN hyperparameter options for side-channel analysis.
\end{enumerate}

The rest of this paper is organized as follows. In Section~\ref{sec:bg}, we provide information about profiling SCA, neural networks, genetic algorithms, and the datasets we use. Section~\ref{sec:related} gives an overview of related works.
In Section~\ref{sec:approach}, we provide details about our novel methodology. Section~\ref{sec:experiments} provides details about the experimental setup and reports the obtained results. In Section~\ref{sec:discussion}, we provide a discussion about the obtained architectures, and finally, in Section~\ref{sec:conclusions}, we conclude the paper.

\section{Background}
\label{sec:bg}
In this section, we cover all necessary background concepts related to Side-channel analysis, neural networks, and genetic algorithms that form the basis of our contribution. The treatment is essential, as a complete overview of these subjects is clearly out of the scope of this manuscript. The reader can find further information in  Picek et al.'s recent systematization of knowledge paper~\cite{picek-sok}.

\subsection{Profiling Side-channel Analysis}
\label{sec:prof-sca}
A \textit{profiling} attack consists of a profiling phase and an attack phase, which are analogous, respectively, to the training and test phase in the context of supervised learning machine learning. In the profiling phase, an attacker uses leakages from a clone device to construct a model that maps the relationship between leakages and corresponding labels (i.e., key-related intermediate data). In the attack phase, he iterates over all possible key candidates and obtains the respective output probabilities for their labels. By repeating this process for each trace and summing the logarithms of the probabilities assigned to each key candidate, he ends up with a log probability vector used to determine the likelihood of each candidate being the correct key. Eq.~\eqref{eq:bg-subkey-candidate-logprob} formulates the procedure of obtaining the log probability for key candidate $k'$ over $N$ attack traces: 
\begin{equation}\label{eq:bg-subkey-candidate-logprob}
    P_{log}(k') = \sum^{N - 1}_{i = 0} \log (P(l(p_i, k'))),
\end{equation}
where $l(\cdot)$ denotes the cryptographic operation that generates the targeted intermediate data, $P(\cdot)$ is the probability assigned by the profiling model, and $p_i$ represents the plaintext used for leakage $i$.

The attack performance is evaluated with the \textit{key rank} metric as follows:
\begin{equation}\label{eq:bg-key-rank}
    \textsf{KR} = \big|\{ k' | P_{\log}(k') > P_{\log}(k) \}\big|.
\end{equation}
Intuitively, the key rank is the number of key candidates with a higher likelihood of correctness than the correct key value. An attack is successful when the correct key is predicted with the highest likelihood or can be brute-forced after being placed among the few highest-likelihood candidates. In this work, we discuss the mean key rank achieved over multiple experimental runs, in which case the metric is commonly referred to as the \textit{guessing entropy}~\cite{10.1007/978-3-642-01001-9_26}. Furthermore, as common in the related works, we will assess the attack performance against a single key byte only (which is denoted partial guessing entropy), but for simplicity, we will denote it as guessing entropy. A common assumption is that attacking a single key byte reveals the average effort required for other key bytes as well~\cite{cryptoeprint:2021/717,Rijsdijk_Wu_Perin_Picek_2021}.

\subsubsection{Countermeasures}
\label{sec:bg-sca-countermeasures}
SCA countermeasures aim to mitigate the information leakage produced during cryptographic operations. 
In this work, we consider the \textit{Boolean masking} and \textit{desynchronization} countermeasures.
The masking countermeasure splits the sensitive intermediate values into different shares to decrease the key dependency~\cite{dpa_book}.
For instance, as implemented in the considered datasets~\cite{ascad-paper}, a random mask $r$ is applied after the \textsf{AddRoundKey} operation in the first round of AES encryption. The intermediate value is then computed as:
\begin{equation}\label{eq:bg-masking}
    Z = \textsf{SBOX}(p \oplus k) \oplus r.
\end{equation}

Desynchronization is a common type of hiding countermeasure introducing time randomness to the leakages. In practice, such effects can be realized by adding clock jitters and inserting random instructions. We simulate this effect by randomly shifting with an upper bound for each trace~\cite{ascad-paper,ae-for-sca-traces}.

\subsection{Neural Networks}
\label{sec:nn}
SCA can be considered a classification task that aims to map the input leakages to a cluster corresponding to the targeted labels. Such a task can be accomplished with a \textit{neural network} (NN), which is essentially a nonlinear function composed of layers of \textit{neurons}, sometimes referred to as \textit{nodes}. The output of a neuron is defined as follows:
\begin{equation}\label{eq:bg-cnns-neuron-func}
    y = \phi( \sum^{n}_{i = i} w_ix_i + b_k),
\end{equation}
which is computed by multiplying the neuron's inputs $x_1, \ldots, x_{n}$ from the previous layer with their corresponding weights $w_1, \ldots, w_{n}$, adding the \textit{bias} value $b_k$ corresponding to neuron $k$, and finally transforming the result with the \textit{activation function} $\phi$. The activation function acts as a source of nonlinearity and often improves the efficiency of the training phase. Two common activation functions for NNs in SCA are the \textit{rectified linear unit} (ReLU) and \textit{scaled exponential linear unit} (SELU). 

When training a neural network, the weight and bias for each neuron are updated with gradient descent to minimize the loss function. A common loss function in multi-class classification problems is the \textit{categorical cross-entropy} (CCE). Cross-entropy is a measure of the difference between two distributions. Minimizing the cross-entropy between the true distribution of the classes and the distribution modeled by the neural network improves its predictions:
\begin{equation}
\label{eq:cce}
    CCE(y,\hat{y})=-\frac{1}{n}\sum^l_{i=1}\sum^c_{j=1}y_{i,j} \cdot \log(\widehat{y_{i,j}}),
\end{equation}
where $c$ and $l$ respectively denote the number of classes and data, $y$ is the true value, and $\hat{y}$ is the predicted value.

A primary type of NN architecture in SCA is the \textit{multilayer perceptron} (MLP), in which a sequence of fully-connected hidden layers of neurons is followed by an output layer that transforms the final output values to label prediction probabilities. A \textit{convolutional neural network} (CNN) is another commonly used type of network in SCA. It prepends its first fully-connected layer with one or more convolutional blocks. Such a block consists of a \textit{convolutional layer} that attempts to compute local features over the input data, and it is optionally followed by a \textit{pooling layer} that aggregates the resulting values, e.g., by calculating $n$-wise averages. Eq.~\eqref{eq:bg-cnns-conv-op} formally displays the application of $j$ convolution filters with kernel size $k$ on inputs $\{x_i, x_{i + 1}, \ldots, x_{i + j}\}$. A convolutional layer repeats such convolutions until shifted through all $n$ inputs, resulting in $n \cdot j$ inputs for the fully-connected layers. Formally, this operation can be stated as follows:
\begin{equation}\label{eq:bg-cnns-conv-op}
\begin{split}
    \textsf{conv}(\{x_i, x_{i + 1}, \ldots, x_{i + j}\}) = \{ & c_{0,0} \cdot c_{0,1} \cdot \ldots \cdot c_{0,k-1} \cdot x_i,\\
    & c_{1,0} \cdot c_{1,1} \cdot \ldots \cdot c_{1,k-1} \cdot x_{i + 1},\\
    & \ldots,\\
    & c_{j-1,0} \cdot c_{j-1,1} \cdot \ldots \cdot c_{j-1,k-1} \cdot x_{i + j} \} .
\end{split}
\end{equation}

\subsection{Genetic Algorithms}
\label{sec:bg-gentic-algorithms}
A genetic algorithm (GA) is a type of population-based optimization algorithm that typically utilizes elements from biological evolution~\cite{10.5555/2810085}. A GA's objective is to optimize a solution to some problem by maintaining a population of such solutions and evolving them over several \textit{generations}. We refer to such a solution as an \textit{individual} or \textit{genome} consisting of building blocks known as \textit{genes}. One generation is performed by evaluating the fitness of each genome, selecting fit genomes as parents for reproduction, and applying genetic operators such as mutation and crossover on those parents to generate the offspring, which represents the next generation.

Before commencing the first generation, the genomes in the population are randomly initialized for diversity. One then starts an iteration of generations until the fitness evaluation budget expires or the fitness value of the best genome achieves a predefined threshold. Each generation starts with \textit{fitness evaluation}, by assigning a fitness value to each genome that measures how well the corresponding individual performs concerning the relative optimization problem. The next step is \textit{selection}, which aims to cull weak genomes from the population so that the algorithm favors genetic modifications that create fitter genomes. Rather than straightforwardly selecting a number of the fittest genomes, modern GAs employ more sophisticated methods to preserve diversity in the population. One such method is \textit{tournament selection}~\cite{Miller95}, which determines parents by holding `tournaments' of some randomly picked genomes and retaining the ones with the best fitness as parents.

Having selected the parents, a GA usually produces as many offspring individuals as the number of parents. Production of one child's genome involves applying one or multiple \textit{genetic operators} to one or multiple parents. These operators include \textit{mutation} and \textit{crossover}~\cite{ga-generic-explanation-holland1992}, though the latter can be omitted in a nonmating GA.
Mutation only requires one parent, cloned and randomly modified with a predefined mutation function to produce one child. On the other hand, crossover refers to the combination of properties of two or more parents to construct a child.
In this work, we apply \textit{polynomial mutation}, a mutation method for real-valued parameters introduced by Deb and Agrawal~\cite{deb-agrawal-polynom-mutation} that is designed for variables with predefined minimum and maximum boundaries. The method mutates some variable $x$ towards either lower boundary $x_L$ or upper boundary $x_U$ with uniform probability. The degree of the mutation is then determined by pseudorandom number $0 \leq u < 1$ and parameter $\eta$, with a higher value of $\eta$ resulting in a smaller mutation range. In other words, the mutated value $x'$ is equal to $x + \bar{\delta}_L (x - x_L)$ or $x + \bar{\delta}_R (x_R - x)$ with $\bar{\delta}_L$ and $\bar{\delta}_R$ scaling with $u$ and $\eta$ as defined in Eq.~\eqref{eq:bg-gas-polynom-mut}.
\begin{equation}\label{eq:bg-gas-polynom-mut}
\begin{split}
    \bar{\delta}_L = & (2u)^{\frac{1}{1 + \eta}} - 1\\
    \bar{\delta}_R = & 1 - (2(1 - u))^{\frac{1}{1 + \eta}}.
\end{split}
\end{equation}

\textit{Neuroevolution} is the usage of an evolutionary algorithm for constructing or optimizing an NN. In this work, we will use a GA to construct a neural network architecture for SCA. In such a scenario, a genome in a GA describes the hyperparameter combination of an NN. The fitness is determined through the evaluation of the network's performance.

\subsection{Datasets}
\label{sec:bg-data-sets}
We use the ASCAD dataset~\cite{ascad-paper}, where each trace comprises 700 trace points corresponding to the S-box operation of the third key byte. Note that we are referring to the \textit{fixed-key} ASCAD dataset, where the same encryption key is used in all AES operations. These traces are protected with the \textit{masking} countermeasure, so the intermediate value $Z$ was computed as in Eq.~\eqref{eq:bg-masking} for some random mask byte $r$.

The training (35\,584) and validation (3\,840) sets are balanced samples taken from the 50\,000 training traces. Their respective numbers were chosen such that both sets are sufficiently large for their respective purposes. Since we use the identity leakage model in all our experiments with this method, both sets' numbers are a multiple of 256, i.e., the number of possible output labels. Note that while we do not expect many issues with the identity leakage model and class imbalance~\cite{Picek_Heuser_Jovic_Bhasin_Regazzoni_2018}, we still balance the classes to mitigate any undesired effects. Ten thousand attack traces are used to assess the attack performance. Finally, we conduct experiments on this dataset with and without the desynchronization countermeasure.

\section{Related Work}
\label{sec:related}
We now give an overview of the literature concerning optimizing NN's architectures for the SCA domain, considering both manual and fully automated approaches. Next, we briefly survey the automated methods based on neuroevolution.

\subsection{Network Architecture Optimization in SCA}
\label{sec:nas-sca}
In 2016, Maghrebi et al. proposed several DL-based approaches and verified their effectiveness on the DPAv2 dataset~\cite{dpav2-dataset} and custom AES implementations with and without first-order masking~\cite{breaking-crypto-with-dl}. The customized neural networks obtained a key rank of zero with fewer than $10^3$ training traces on the masked implementation. Interestingly, their CNN architecture was determined with a genetic algorithm using the guessing entropy as a fitness function, but they do not provide in-depth elaboration on their methodology.

Benadjila et al.~\cite{ascad-paper} further explored the performances of neural networks in the SCA context. Upon evaluating several promising architectures from the existing literature, VGG-16~\cite{vgg-16-competition-paper} or deep CNNs with similar structures were deemed well-suited for SCA. 
Kim et al.~\cite{noise-cnss-sca} enhanced the attack performance in combination with data augmentation, which turned out to be an effective way of priming the CNN to deal with several kinds of countermeasures. Next, Zaid et al. introduced efficient CNN architectures~\cite{efficient-cnn-arch-method} that could obtain state-of-the-art performance with significantly reduced neural network size. Wouters et al.~\cite{revisiting-efficient-cnn-archs} further reduced the networks' size with data preprocessing strategies.

Besides manually optimizing the neural network, recent research has also attempted fully automated approaches for network architecture search. Rijsdijk et al.~\cite{Rijsdijk_Wu_Perin_Picek_2021} customized the MetaQNN reinforcement learning algorithm for SCA to automatically find CNN architectures. However, the search space is roughly limited to hyperparameters that we know to be effective, and pure MLP architectures are not discussed. Each NN is evaluated by training it for 50 epochs using the Adam optimizer and the SELU activation function in their work. Wu et al. proposed AutoSCA~\cite{lichao-auto-hyperparam-tuning}, which uses \textit{Bayesian optimization} to find architecture hyperparameters for both MLPs and CNNs. Their approach produced good results and mainly focused on finding larger architectures with at least 100 neurons in each dense layer.

The application of neuroevolution to perform side-channel analysis has only scarcely been explored in existing work. Knezevic et al. used genetic programming to evolve custom activation functions specific to side-channel analysis~\cite{knezevic-neurosca-gp} that can outperform the widely used ReLU function. The genome in their approach encodes an activation function as a tree containing unary and binary operators, with leaves representing the function's inputs. Such a tree is initialized with a depth of two to five levels and limited to twelve levels during evolution. For fitness evaluation, they compute the mean number of attack traces required to obtain a key rank of zero over one hundredfold and add it to one minus the accuracy. 
The method resulted in novel activation functions that improve performance on large and efficient MLPs and CNNs. Acharya et al. proposed InfoNEAT~\cite{infoneat}, an approach that tailors the \textit{NeuroEvolution of Augmenting Topologies} (NEAT) algorithm~\cite{neat-paper} specifically for side-channel analysis. Their approach considers the identity leakage model and uses NEAT to evolve an NN architecture with a single output node for each of the 256 output classes. The resulting 256 binary networks are combined by a \textit{stacking} approach that uses the networks' outputs as inputs for a logistic regression model. Such a stacked model is created for multiple different folds of balanced trances taken from the complete dataset, after which those models' prediction probabilities can be summed to form a final prediction for each attack trace.

\subsection{Evolution-based Network Architecture Search}
\label{sec:evo-nas-sca}
Evolutionary approaches have been widely used in automated network architecture searches. Real et al. developed one such method for image classification on modern datasets~\cite{large-scale-evo-of-img-clf-2017}, a task that requires large networks. They propose a nonmating GA with both NEAT-like~\cite{neat-paper} mutations and layer-level mutations to evolve CNNs on granular and large scales. Specifically, each genome is trained with backpropagation on 45\,000 samples before evaluating its fitness. In this approach, a child's genome keeps the weights and biases of its parent, effectively training each network over time.

Other successful neuroevolution methods that construct CNNs for image classification include the Deep Evolutionary Network Structured Representation approach DENSER~\cite{denser-cnn-neuroevo} and EvoCNN~\cite{cnn-evo-for-clf-2019}. DENSER uses a 2-level genotype where the first level encodes the NN's hyperparameters while the second encodes layer-specific variables such as the number of neurons or the variables for a convolutional filter. This structure enables the algorithm to be used for MLPs, CNNs, and other types as long as they can be appropriately defined in the genome's second level. Furthermore, DENSER trains NNs with backpropagation before evaluating the fitness on a validation set. EvoCNN works similarly but evolves the weight initialization values along with the architecture hyperparameters.

\section{NASCTY}
\label{sec:approach}
\textit{Neuroevolution to Attack Side-channel Traces Yielding Convolutional Neural Networks} (NASCTY-CNNs) is a GA that modifies hyperparameters of CNNs for side-channel analysis. Algorithm~\ref{alg:nascty} shows the main procedure of our approach. This section will specify the genome structure, the initialization of the population, the fitness evaluation method, and the method used to produce offspring.
\begin{algorithm}[t]
    \caption{The NASCTY-CNNs algorithm
    \label{alg:nascty}}
    \begin{algorithmic}[1]
        \Statex
        \Let{$train\_traces, train\_labels, valid\_traces, valid\_labels$}{\textsf{sample($ascad\_data$)}}
        \Let{$pop$}{\textsf{initialise\_population()}}
        
        \While{$gen < max\_gens$}
        
        \State \textsf{evaluate\_fitness\_values($pop, train\_traces, train\_labels, train_plaintexts$)}
        \Let{$parents$}{\textsf{tournament\_selection($pop$)}}
        \Let{$offspring$}{\textsf{produce\_offspring($parents$)}}
        \Let{$pop$}{$parents \cup offspring$}
        \EndWhile
        
        \State \Return{genome in $pop$ with the lowest fitness}
    \end{algorithmic}
\end{algorithm}
Note that the sampling of training and validation data is only performed once, meaning that we use the same data for fitness evaluation in every generation. Furthermore, we only use balanced data samples, i.e., a sampled set of traces is taken such that it contains an equal number of traces corresponding to each possible output label. Following~\cite{large-scale-evo-of-img-clf-2017,denser-cnn-neuroevo,cnn-evo-for-clf-2019}, we also use tournament size 3 during selection, which means we randomly choose three individuals and select the fittest one as a potential parent for reproduction. Finally, note that all experiments are performed targeting the third (masked) key byte of the fixed-key ASCAD dataset in which each trace consists of 700 trace points, each of which is normalized between -1 and 1.

\subsection{Genome Structure}
\label{sec:gen-struct}
\begin{figure}[b]
    \captionsetup[subfigure]{justification=centering}
    \centering
    \includegraphics[width=.75\textwidth]{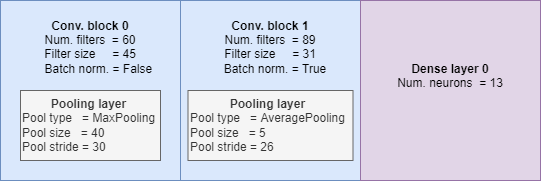}
    \caption{An example of the genome encoding used in the NASCTY algorithm.}
    \label{fig:nascty-genome-example}
\end{figure}

The NASCTY genome represents a CNN and consists of a list of zero up to and including five convolutional blocks, an optional pooling layer when no convolutional blocks are present, and a list of one up to and including five dense layers. Each of the convolutional blocks is described with the number of convolutional filters, the filter size, a Boolean denoting the presence of a batch normalization layer, and a pooling layer. Any pooling layer in the genome comprises a pooling type, either max pooling or average pooling, a pool size, and a pool stride. Finally, a dense layer is described only by its number of neurons. An example of the genome structure is presented in Figure~\ref{fig:nascty-genome-example}.

When expressing a NASCTY genome as a neural network, following the state-of-the-art architectures, we always use the SELU activation function for all hidden neurons, use He weight initialization for the convolutional blocks and dense layers, and use Glorot uniform weight initialization for the output layer, which uses the softmax activation function. Although enlarging the genome parameter spaces would increase the diversity of the populations, applying the prior knowledge would speed up the evolution process. 

\subsection{Population Initialization}
\label{sec:pop-init}
We initialize all networks randomly with hyperparameters within the ranges displayed in Table~\ref{tab:nascty-genome-param-ranges}. These hyperparameter ranges, as well as the genome structure itself, are inspired by the VGG-like networks for SCA found in prior work~\cite{ascad-paper,noise-cnss-sca}, as well as by the automated architecture search approach for SCA based on reinforcement learning~\cite{Rijsdijk_Wu_Perin_Picek_2021}. 
\begin{table}[t]
\small
    \centering
    \begin{tabular}{lc}
    \toprule
    \textbf{Parameter}    & \textbf{Options}     \\ 
    \midrule
    Num. convolutional blocks  & 0 to 5 in a step of 1   \\ 
    \midrule
    Num. dense layers          & 1 to 5 in a step of 1   \\ 
    \midrule
    Num. convolutional filters & 2 to 128 in a step of 1 \\ 
    \midrule
    Filter size                & 1 to 50 in a step of 1  \\ 
    \midrule
    Batch normalisation layer  & False, True    \\ 
    \midrule
    Pooling type               & Average, Max   \\ 
    \midrule
    Pool size                  & 2 to 50 in a step of 1  \\ 
    \midrule
    Pool stride                & 2 to 50 in a step of 1  \\ 
    \midrule
    Num. dense neurons         & 1 to 20 in a step of 1  \\ 
    \bottomrule
    \end{tabular}
    \caption{Ranges for CNN genome hyperparameters in the NASCTY algorithm.}
    \label{tab:nascty-genome-param-ranges}
\end{table}

Note that one can initialize the population by starting with architectures with a minimal number of trainable parameters to reduce the required evaluation time. However, we opt to completely initialize the population at random to avoid local optima that may come about due to the reduced diversity in the population.

\subsection{Fitness Evaluation}
\label{sec:fit-eval}
Once a genome is defined, the corresponding CNN is trained with the Adam optimizer.\footnote{All networks are trained with the same seed in every generation to ensure they are fairly compared.} The loss value on the validation set is used for the fitness evaluation. By minimizing loss, we aim to have a system aligned with the related works in DL-SCA. Naturally, one could consider other options here, for instance, the ones applied in~\cite{lichao-auto-hyperparam-tuning}. The objective of training the networks before evaluating them is to enable us to differentiate their quality more accurately. We chose to train each network for ten epochs as other works do~\cite{lichao-auto-hyperparam-tuning,denser-cnn-neuroevo,cnn-evo-for-clf-2019} and preliminary experiments following the methodology recommendations by~\cite{cnn-evo-for-clf-2019} showed that this is enough for similarly sized networks to observe significant CCE differences in networks of different qualities.

\subsection{Offspring Production}
\label{sec:off-prod}
After evaluating the fitness of each genome in one generation, the members of the next generation (offspring) can be produced. Half of these members are produced by applying tournament selection to the population to find fit genomes that will act as parents. The remaining half, on the other hand, is constructed by randomly choosing pairs of those parents on which the crossover and mutation operations are applied. Optionally, we may only apply tournament selection to some proportion of the top-performing members of the population to select parents. This operation is performed to ensure the best genomes are maintained in the population, a concept known as elitism which can be tuned with a \textit{truncation proportion} parameter.  

Our algorithm uses one of two possible types of crossover, i.e., either \textit{one-point} crossover or \textit{parameter-wise} crossover, both of which are common crossover strategies in genetic algorithms. The performance of these two methods is evaluated in Section~\ref{sec:nascty-results-grid-search}. To enact a one-point crossover with two parents, we apply one-point crossover separately on the parents' lists of convolutional blocks and dense layers. For a list of either convolutional blocks or dense layers, we achieve this operation by picking a random cutoff point in both parents' lists of that layer type. The first child's list of that type is then created by connecting the first parent's list before its cutoff point to the second parent's list after its cutoff point, while the second child's list is created by connecting the remaining units. The one-point crossover operation is then finalized by randomly dividing the parents' optional pooling layers that are present in the absence of convolutional blocks among the offspring.

\begin{figure}[H]
    \captionsetup[subfigure]{justification=centering}
    \centering
        \begin{subfigure}[t]{0.45\paperwidth}
            \centering
            \includegraphics[width=0.9\textwidth]{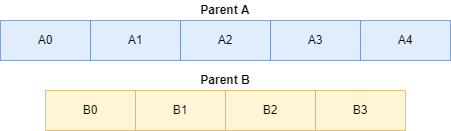}
            \caption{Each parent's list of layers or blocks.}
            \label{fig:nascty-onepoint-co-1}
        \end{subfigure}
        
        \phantom{M}
        
        \begin{subfigure}[t]{0.45\paperwidth}
            \centering
            \includegraphics[width=0.9\textwidth]{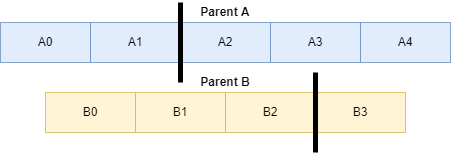}
            \caption{The parents' respective random cutoff points.}
            \label{fig:nascty-onepoint-co-2}
        \end{subfigure}

        \phantom{M}
        
        \begin{subfigure}[t]{0.5\paperwidth}
            \centering
            \includegraphics[width=0.9\textwidth]{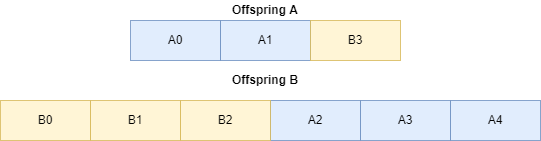}
            \caption{The two lists of layers or blocks present in the offspring after crossover.}
            \label{fig:nascty-onepoint-co-3}
        \end{subfigure}
    \caption{Visualisation of one-point crossover on a list of convolutional blocks or dense layers in a NASCTY genome.}
    \label{fig:nascty-onepoint-co}
\end{figure}

In our implementation of parameter-wise crossover, the first child genome is created by iterating over the parents' pairs of convolutional blocks and randomly inheriting convolutional block genes from either parent. The second child's genome then inherits the remaining hyperparameters for these blocks. The exact process is repeated for the parents' lists of dense layers. In the typical scenario where one parent has more convolutional blocks or dense layers than the other parent, the excess units are appended unmodified to the first child's genome.

After the crossover operation, the offspring are mutated through one of the following methods with a uniform probability:
\begin{itemize}
    \item Adding one random convolutional block or dense layer with randomly initialized hyperparameters;
    \item Removing one random convolutional block or dense layer;
    \item Modifying all hyperparameters through \textit{polynomial mutation} with probability $\frac{1}{n}$, where $n$ is the total number of modifiable hyperparameters in the genome.
\end{itemize}
The polynomial mutation is described in Section~\ref{sec:bg-gentic-algorithms} and is designed for variables with predefined minimum and maximum boundaries, which fits our task of exploring proper hyperparameter values within predefined ranges. This mutation method comes with a parameter $\eta$, which can be increased to reduce the degree to which a given parameter is modified. The value for $\eta$ is recommended to be between 20 and 100~\cite{deb-mut-scheme-analysis}, though experimentation is the only way to determine a proper value of $\eta$ for an untested scenario.

\section{Experiments}
\label{sec:experiments}
In this section, we discuss the experimental evaluation of NASCTY. We start by describing the setup of our experiments. Then, we show the outcome of the preliminary tuning phase based on grid search. Finally, we present the results obtained by GA with the best-performing parameter combination on masked and desynchronized traces of ASCAD.

\subsection{Experimental Setup}
\label{sec:exp-setup}
For the experimental validation of our approach, we optimize GA parameters through a grid search, then evaluate the performance on the masked ASCAD traces and masked and desynchronized ASCAD traces for several desynchronization levels. The objective of these experiments is to determine:
\begin{itemize}
    \item The effectiveness of GA parameters for our approach;
    \item Whether our automated approach can produce NNs that outperform similar NNs found through trial and error;
    \item Architecture components that contribute to the effectiveness of an SCA.
\end{itemize}
To account for the randomness introduced by the mutation operations, we run five experiments for each GA parameter configuration and report the best results. The best genome resulting from the NASCTY algorithm is evaluated by training its corresponding NN for 50 epochs and computing the mean incremental key rank~\cite{Rijsdijk_Wu_Perin_Picek_2021} over 100 folds.

All experiments are executed with 52 parallel workers, each of which runs at approximately 2.1GHz on an Intel E5-2683 v4 CPU. With these computational resources, the discussed experiments required 84GB RAM and took at least four and at most seven days to complete. This significant variance in runtime complexity is caused by the pseudorandom nature of GAs, which results in the construction and evaluation of NNs of varying sizes.

\subsection{Parameter Tuning by Grid Search}
\label{sec:nascty-results-grid-search}
Table~\ref{tab:nascty-grid-search-params} shows a summary of the GA parameters for the grid search. Note that additional parameter options, mutation strategies, and crossover strategies could potentially result in better performance, but such adjustments would have to significantly diverge from our current strategy to assess the general effectiveness of the algorithm. We run each grid search experiment with a population size of 52 to match the number of available parallel workers and run the GA for ten generations. Furthermore, each of these experiments uses the same training data, validation data, and initial population to observe the impact of the parameter changes more accurately.
\begin{table}[t]
\footnotesize
    \centering
    \begin{tabular}{lc}
    \toprule
    \textbf{Parameter} & \textbf{Options}                 \\ \midrule
    Polynomial mutation $\eta$ & 20, 40                   \\ \midrule
    Crossover type          & One-point, parameter-wise \\ \midrule
    Truncation proportion   & 0.5, 1.0                   \\ \bottomrule
    \end{tabular}
    \caption{The parameter values we consider during grid search for NASCTY.}
    \label{tab:nascty-grid-search-params}
\end{table}

\begin{figure}[t]
    \captionsetup[subfigure]{justification=centering}
    \centering
        \begin{subfigure}[t]{0.5\textwidth}
            \centering
            \includegraphics[width=\textwidth]{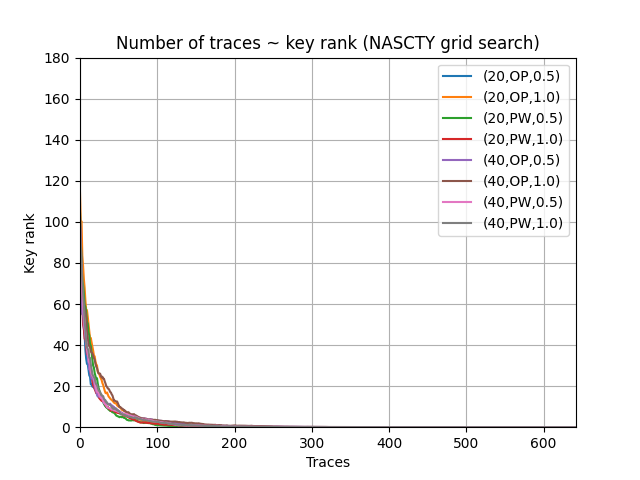}
            \caption{Best runs' mean performance over 100 folds}
            \label{fig:nascty-grid-search-traces-vs-keyrank}
        \end{subfigure}%
        \begin{subfigure}[t]{0.5\textwidth}
            \centering
            \includegraphics[width=\textwidth]{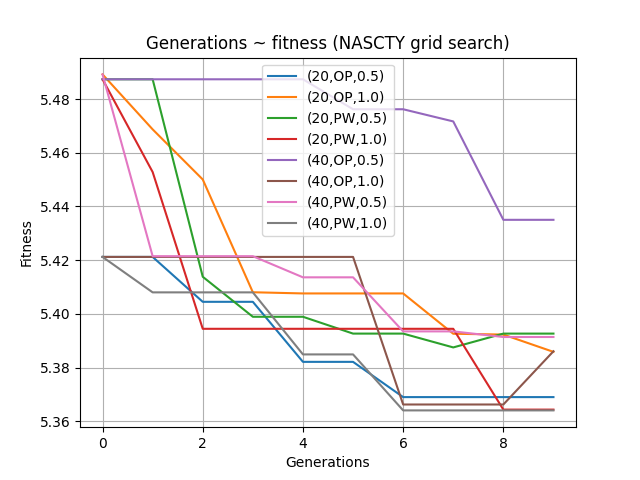}
            \caption{Best runs' fitness progressions (the lower the better)}
            \label{fig:nascty-grid-search-gens-vs-fitness}
        \end{subfigure}
    \caption{NASCTY grid search results corresponding to the best network obtained with each parameter combination.}
    \label{fig:nascty-grid-search}
\end{figure}

We expect lower values of the polynomial mutation parameter $\eta$~\cite{deb-agrawal-polynom-mutation} a better choice in these experiments where we run for relatively few generations. More specifically, smaller values of $\eta$ cause larger mutations, which carry the potential to find better networks faster, but a value of 20 for $\eta$ is still large enough to avoid unreasonably risky mutations~\cite{deb-mut-scheme-analysis}. For similar reasons, a larger truncation proportion would result in better performance as it preserves diversity in an already-small population. However, its influence is likely not as significant as that of the chosen crossover and mutation configurations since those can modify the population more straightforwardly. Finally, we expect either crossover strategy to perform well since both allow the algorithm to find effective architectures in the predefined hyperparameter ranges. The performance of the final network of each parameter combination's best run is shown in Figure~\ref{fig:nascty-grid-search}.

Figure~\ref{fig:nascty-grid-search-traces-vs-keyrank} implies that each parameter combination is capable of producing fit architectures for the considered ASCAD traces. Similarly, the best runs' fitness plots over generations shown in Figure~\ref{fig:nascty-grid-search-gens-vs-fitness} demonstrate that the best genome's CCE, i.e., the validation loss, can improve significantly in as few as ten generations, regardless of the parameter combination under consideration.

The best mean incremental key rank among all grid search experiments was approximately 0.50419 and resulted from the experiments with a polynomial mutation $\eta$ value of 20, one-point crossover, and a truncation proportion of 1.0. Therefore, those parameters are applied for all further experiments. To determine each GA parameter's influence on the final performance, we observe the effect of modifying one variable at a time while keeping the others constant at the aforementioned best-observed values. Table~\ref{tab:nascty-grid-search-param-influence} displays how such modifications affect the mean incremental key rank. From the table, we can infer that only the crossover strategy significantly affects the final performance among the parameters we considered; one-point crossover is preferred over the parameter-wise crossover. Since the best runs using parameter-wise crossover in Figure~\ref{fig:nascty-grid-search} still perform well, the performance difference likely results from poor consistency compared to runs using the one-point crossover. We suspect that the additional consistency observed with one-point crossover is achieved through its advantage in retaining functional sequences of convolutional blocks or dense layers. In addition, one-point crossover on lists of layers intuitively provides synergy with our mutation strategy of adding or removing an entire layer because effective additions or removals can be identified more quickly when they are separated into offspring in a modular fashion.

\begin{table}[b]
\footnotesize
    \centering
    \begin{tabular}{lccc}
    \hline
    \textbf{$\eta$} & \textbf{Crossover type} & \textbf{Truncation proportion} & \textbf{Mean incremental key rank} \\ \midrule
    20                               & One-point               & 1.0                            & 0.50419                       \\ \midrule
    40                               & One-point               & 1.0                            & 0.50880                       \\ \midrule
    20                               & Parameter-wise          & 1.0                            & 0.89354                       \\ \midrule
    20                               & One-point               & 0.5                            & 0.50966                       \\ \bottomrule
    \end{tabular}
    \caption{The effect of each parameter on the final incremental key rank in NASCTY grid search experiments.}
    \label{tab:nascty-grid-search-param-influence}
\end{table}

\subsection{ASCAD: Masked and Desynchronized}
\label{sec:ascad-res}
All remaining experiments are run with the best-performing parameter options found through our grid search experiments. In addition, we run these experiments with a population size of 100 to fully exploit the resources at our disposal. In contrast to our grid search experiments, these experiments do not use a seed for the pseudorandom numbers involved anywhere in the GA except for the fitness evaluation procedure, in which we use a seed for the training of each NN to ensure the genomes are fairly compared.

We first run NASCTY on masked ASCAD traces for 75 generations to evaluate the algorithm's general effectiveness. Then, we run experiments on the same masked dataset, further protected with desynchronization as described in Section~\ref{sec:bg-sca-countermeasures}. Specifically, we run three sets of experiments with desynchronization levels of 10, 30, and 50, respectively. With this approach, we aim to determine whether NASCTY can circumvent or mitigate countermeasures without additional algorithm modifications and whether larger desynchronization levels hinder NASCTY's ability to find good architectures.

\begin{figure}[t]
    \captionsetup[subfigure]{justification=centering}
    \centering
        \begin{subfigure}[t]{0.5\textwidth}
            \centering
            \includegraphics[width=\textwidth]{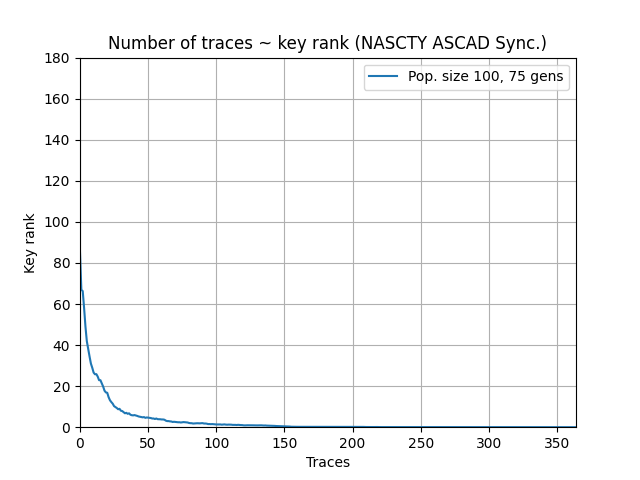}
            \caption{Best NN's mean performance over 100 folds}
            \label{fig:nascty-ascasd-masked-traces-vs-keyrank}
        \end{subfigure}%
        \begin{subfigure}[t]{0.5\textwidth}
            \centering
            \includegraphics[width=\textwidth]{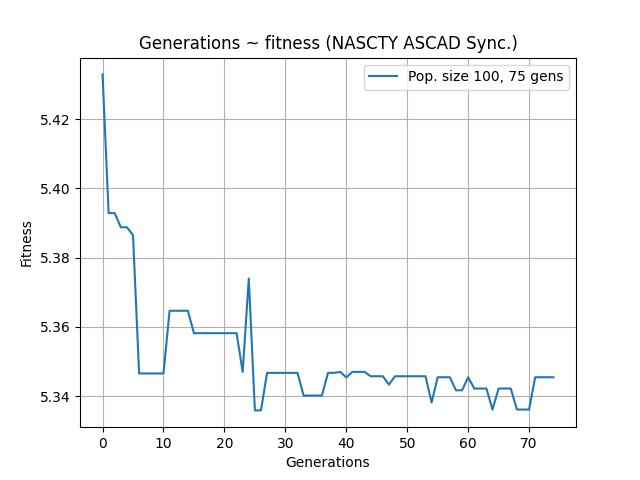}
            \caption{Best run's fitness progression}
            \label{fig:nascty-ascad-masked-gens-vs-fitness}
        \end{subfigure}
    \caption{NASCTY results corresponding to the best network obtained on masked ASCAD traces.}
    \label{fig:nascty-ascad-masked}
\end{figure}

The fitness progression trends in Figure~\ref{fig:nascty-grid-search-gens-vs-fitness} indicate that more generations would improve the observed fitness of the best genome. Following this, we first ran NASCTY on masked ASCAD traces for 75 generations with a population size of 100.
Figure~\ref{fig:nascty-ascad-masked} shows the results of this first experiment.
As shown in Figure~\ref{fig:nascty-ascasd-masked-traces-vs-keyrank}, the best-obtained network converges smoothly. Ultimately, the network breaks the target in 314 attack traces and achieves a mean incremental key rank of 0.51857. The best fitness value progresses (Figure\ref{fig:nascty-ascad-masked-gens-vs-fitness}) continually after ten generations have passed, then stagnates well before the seventy-fifth generation is reached. Moreover, despite the difference in the population size and the number of generations, the best network is outperformed by several of the NNs obtained with our grid search experiments. This observed fitness stagnation implies that the algorithm may be prone to get stuck in local optima. Typically, the mutation is the source of global search in a GA, so we recommend that future work evaluates lower values of $\eta$ for polynomial mutation or possibly more perturbing mutation strategies.

\begin{figure}[t]
    \captionsetup[subfigure]{justification=centering}
    \centering
        \begin{subfigure}[t]{0.5\textwidth}
            \centering
            \includegraphics[width=\textwidth]{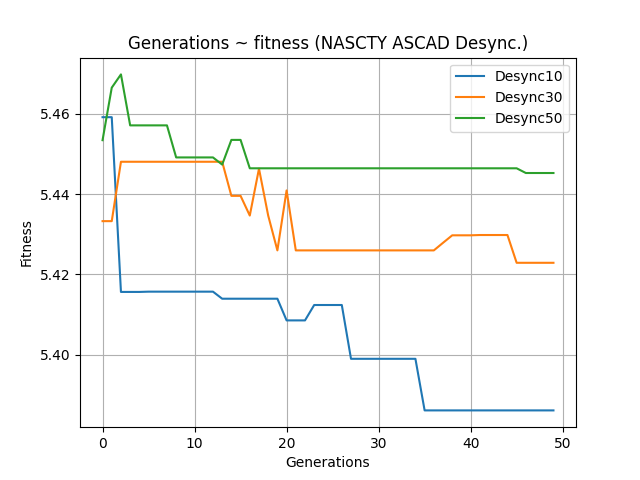}
            \caption{Best runs' fitness progressions}
            \label{fig:nascty-ascad-desync-fitprog}
        \end{subfigure}%
        \begin{subfigure}[t]{0.5\textwidth}
            \centering
            \includegraphics[width=\textwidth]{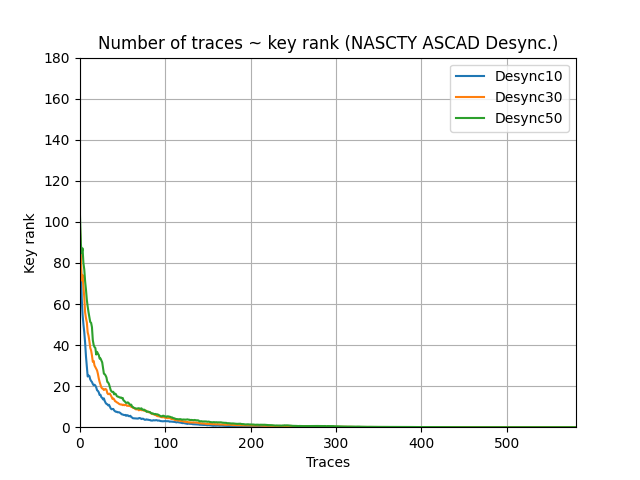}
            \caption{Best runs' mean performance over 100 folds}
            \label{fig:nascty-ascad-desync-traces-vs-keyrank}
        \end{subfigure}
    \caption{NASCTY results for ASCAD traces at desynchronization levels 10, 30, and 50.}
    \label{fig:nascty-ascad-desync}
\end{figure}

Due to the observation of fitness progress in the previous experiment, we ran NASCTY on the masked and desynchronized ASCAD traces for 50 generations instead of 75. The results for desynchronization levels 10, 30, and 50 are displayed in Figure~\ref{fig:nascty-ascad-desync}.
As shown in Figure~\ref{fig:nascty-ascad-desync-fitprog}, the time-randomness introduced by desynchronization affects the algorithm's performance, considering both fitness progress and final performance are noticeably diminished as the desynchronization level increases.
Still, the results show that NASCTY can find effective architectures despite the added countermeasures, with the networks evaluated in Figure~\ref{fig:nascty-ascad-desync-traces-vs-keyrank} being able to obtain key rank 0 in 338, 474, or 531 traces respectively for desynchronization levels 10, 30, and 50.

\section{Discussion}
\label{sec:discussion}
\begin{figure}[t]
    \captionsetup[subfigure]{justification=centering}
    \centering
        \begin{subfigure}[t]{0.24\textwidth}
            \centering
            \includegraphics[width=\textwidth]{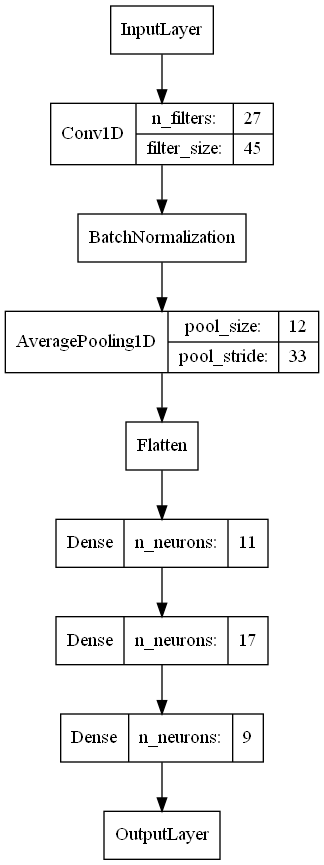}
            \caption{Synchronized}
            \label{fig:nascty-model-visualisations-sync}
        \end{subfigure}%
        \begin{subfigure}[t]{0.24\textwidth}
            \centering
            \includegraphics[width=\textwidth]{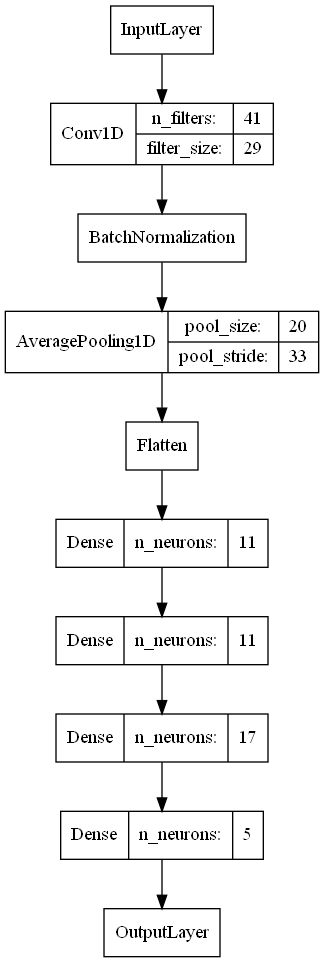}
            \caption{Desynchronization level 10}
            \label{fig:nascty-model-visualisations-desync10}
        \end{subfigure}
        \begin{subfigure}[t]{0.24\textwidth}
            \centering
            \includegraphics[width=\textwidth]{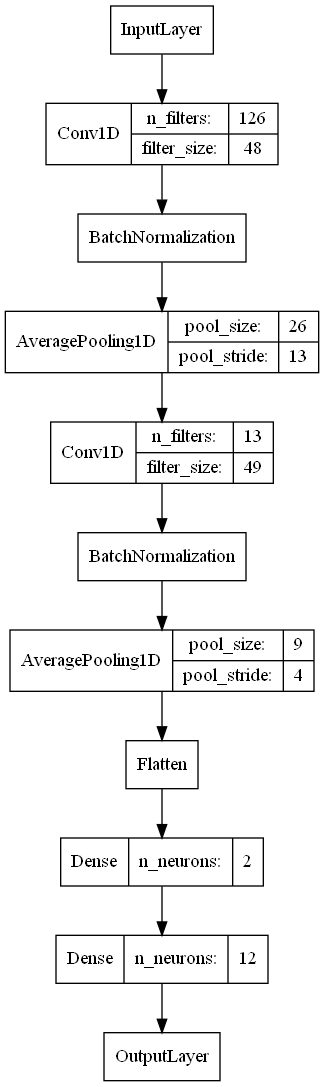}
            \caption{Desynchronization level 30}
            \label{fig:nascty-model-visualisations-desync30}
        \end{subfigure}%
        \begin{subfigure}[t]{0.24\textwidth}
            \centering
            \includegraphics[width=\textwidth]{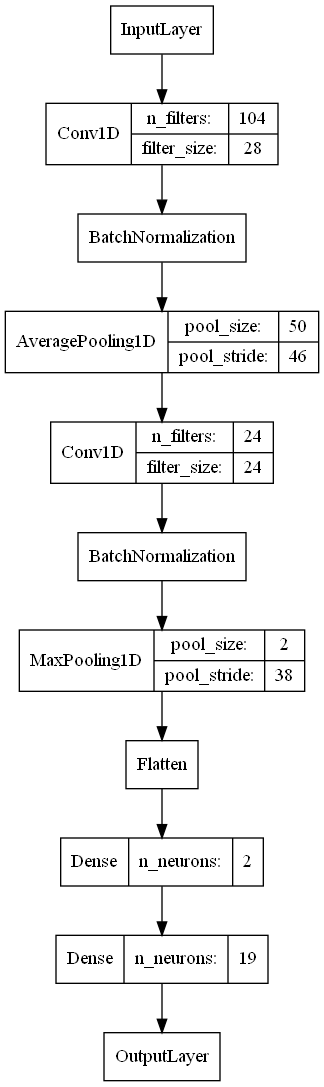}
            \caption{Desynchronization level 50}
            \label{fig:nascty-model-visualisations-desync50}
        \end{subfigure}
    \caption{Best architectures produced by NASCTY for masked ASCAD traces protected with several desynchronization levels.}
    \label{fig:nascty-model-visualisations}
\end{figure}

The best run on the synchronized ASCAD traces produced the CNN architecture shown in Figure~\ref{fig:nascty-model-visualisations-sync}. It has 10\,470 trainable parameters and vaguely resembles the efficient CNN proposed by Zaid et al.~\cite{efficient-cnn-arch-method}.
In comparison, the architecture produced with NASCTY has an additional dense layer and possesses several unintuitive components, such as 27 convolutional filters of size 45 and a pool stride that exceeds the pool size. Since the efficient MLP proposed by Wouters et al.~\cite{revisiting-efficient-cnn-archs} only required two layers of ten neurons each, we surmise that NASCTY may be inclined to include unnecessary model complexity. In other words, NASCTY does not sufficiently discourage redundant model complexity. Indeed, by increasing the desynchronization from 30 to 50, the number of trainable parameters of the best architecture decreases from 90\,379 to 68\,427. A possible solution would be to introduce a model size penalty in the fitness function. Still, NASCTY, with the current configuration, is sufficient in generating good network architectures. Table~\ref{tab:nascty-sota-comparison} shows that NASCTY can compete with other state-of-the-art automated hyperparameter tuning methods for SCA, although it is ultimately outperformed.
\begin{table}[b]
\small
    \centering
    \begin{tabular}{l|c|c}
    \toprule
    \textbf{Method}    & \textbf{Num. traces to} & \textbf{Num. trainable parameters}    \\ 
    & \textbf{obtain mean key rank 0} &  \\ 
    \midrule
    RL-SCA~\cite{Rijsdijk_Wu_Perin_Picek_2021}  & 242 & 1\,282   \\ 
    \midrule
    AutoSCA~\cite{lichao-auto-hyperparam-tuning} & 158 & 54\,752   \\ 
    \midrule
    NASCTY & 314 & 10\,470 \\ 
    \bottomrule
    \end{tabular}
    \caption{Comparison of automated hyperparameter tuning methods for SCA on synchronized, masked ASCAD traces.}
    \label{tab:nascty-sota-comparison}
\end{table}

The architecture corresponding to a desynchronization level of 10 is similar to the architecture for synchronized traces in both structure and size, with the main difference being the addition of more convolutional filters and another dense layer. Both architectures also feature a pooling layer with a stride that exceeds its size, suggesting their convolutional layers have produced features that are either redundant or incorrectly utilized.

The NASCTY architectures for the two more severe desynchronization levels (30 and 50) provide more insight into NASCTY's way of mitigating this countermeasure. As can be seen in Figures~\ref{fig:nascty-model-visualisations-desync30} and~\ref{fig:nascty-model-visualisations-desync50}, both of these architectures start with a convolutional layer with over 100 filters, a significantly larger number than that of the other architectures' convolutional layers. In addition, both feature two convolutional blocks and two dense layers, with the first dense layer in each network having two neurons. The larger number of filters is consistent with existing approaches to mitigate desynchronization, but the usage of such small dense layers is uncommon when attempting to break protected ASCAD traces. Finally, average pooling appears to be the preferred pooling type in these networks, with max pooling only occurring once. We emphasize that our architecture for desynchronization of 50 outperforms the approach of~\cite{Rijsdijk_Wu_Perin_Picek_2021} or performs very similarly to the state-of-the-art methods proposed in~\cite{efficient-cnn-arch-method,revisiting-efficient-cnn-archs}.

\section{Conclusions and Future Work}
\label{sec:conclusions}
This paper proposed a genetic algorithm for network architecture search in the SCA domain. In NASCTY, each genome encodes the hyperparameters representing a CNN's architecture, and a genome's fitness is evaluated by the validation loss. During offspring production, we apply either one-point crossover on parents' lists of layers or parameter-wise crossover to create a pair of child genomes that we immediately mutate by adding a layer, removing a layer, or applying polynomial mutation on its genes. With this approach, NASCTY could produce comparable architectures to state-of-the-art techniques. The redundant complexity and unintuitive architecture components found in some NASCTY networks suggest that our method can likely be improved further, implying unexplored potential to match or surpass current state-of-the-art approaches.

Furthermore, NASCTY found effective architectures for traces protected with masking and desynchronization levels up to 50 while keeping its GA parameters and implementation largely unmodified. However, the desynchronization did affect the final networks' performances: networks produced by NASCTY for desynchronization levels 0, 10, 30, and 50 obtained key rank 0 within 314, 338, 474, and 531 attack traces, respectively. We recommend that future work evaluate NASCTY's effectiveness and resulting architecture patterns on traces protected with other countermeasures. The observed network architectures showed that NASCTY tends to combat desynchronization by adding a convolutional layer and increasing the number of filters in the first convolutional layer, which was the case in architectures generated for desynchronization levels 30 and 50. Interestingly, these networks started their fully-connected part with a dense layer of two neurons. Additionally, some architectures contained pooling layers of which the stride was larger than their size, which is uncommon in other approaches and suggests that NASCTY may be generating redundant features through unnecessarily large numbers of convolutional filters. The architectures NASCTY generated for synchronized and mildly desynchronized traces also came with more trainable parameters than models from related work, that achieve better performance on the same task~\cite{revisiting-efficient-cnn-archs}, corroborating the hypothesis that NASCTY is prone to adding redundant model complexity. Regardless of the presence of desynchronization, average pooling was preferred to max-pooling in nearly all pooling layers.

For future work, it would be interesting to explore the application of a complexity penalty to fitness evaluation or population initialization with minimal architectures to reduce the network size. Furthermore, we recommend experimenting with more mutation parameters to stimulate a better global search to avoid getting stuck in a local optimum, which often seems to occur well before the algorithm terminates. Once those drawbacks have been resolved, we may find better, more interesting architectures by expanding the search space. To do so, we suggest introducing new hyperparameters to the genome, e.g., activation functions for each layer and the learning rate for more general applications. Finally, our approach considers the ASCAD dataset with a fixed key. It will be interesting to see how well our approach works for other, more difficult datasets, like ASCAD with random keys.

\bibliographystyle{abbrv}
\bibliography{bibliography}

\begin{thebibliography}{10}

\bibitem{infoneat}
R.~Y. Acharya, F.~Ganji, and D.~Forte.
\newblock Infoneat: Information theory-based neuroevolution of augmenting
  topologies for side-channel analysis.
\newblock {\em arXiv preprint arXiv:2105.00117}, 2021.

\bibitem{denser-cnn-neuroevo}
F.~Assun{\c{c}}{\~a}o, N.~Louren{\c{c}}o, P.~Machado, and B.~Ribeiro.
\newblock Denser: deep evolutionary network structured representation.
\newblock {\em Genetic Programming and Evolvable Machines}, 20(1):5--35, 2019.

\bibitem{ascad-paper}
R.~Benadjila, E.~Prouff, R.~Strullu, E.~Cagli, and C.~Dumas.
\newblock Study of deep learning techniques for side-channel analysis and
  introduction to ascad database.
\newblock {\em ANSSI, France \& CEA, LETI, MINATEC Campus, France.}, 22:2018,
  2018.

\bibitem{deb-agrawal-polynom-mutation}
K.~Deb and S.~Agrawal.
\newblock A niched-penalty approach for constraint handling in genetic
  algorithms.
\newblock In {\em Artificial Neural Nets and Genetic Algorithms}, pages
  235--243. Springer, 1999.

\bibitem{deb-mut-scheme-analysis}
K.~Deb and D.~Deb.
\newblock Analysing mutation schemes for real-parameter genetic algorithms.
\newblock {\em International Journal of Artificial Intelligence and Soft
  Computing}, 4(1):1--28, 2014.

\bibitem{10.5555/2810085}
A.~E. Eiben and J.~E. Smith.
\newblock {\em Introduction to Evolutionary Computing}.
\newblock Springer Publishing Company, Incorporated, 2nd edition, 2015.

\bibitem{ga-generic-explanation-holland1992}
J.~H. Holland.
\newblock Genetic algorithms.
\newblock {\em Scientific american}, 267(1):66--73, 1992.

\bibitem{noise-cnss-sca}
J.~Kim, S.~Picek, A.~Heuser, S.~Bhasin, and A.~Hanjalic.
\newblock Make some noise. unleashing the power of convolutional neural
  networks for profiled side-channel analysis.
\newblock {\em IACR Transactions on Cryptographic Hardware and Embedded
  Systems}, pages 148--179, 2019.

\bibitem{knezevic-neurosca-gp}
K.~Knezevic, J.~Fulir, D.~Jakobovic, and S.~Picek.
\newblock Neurosca: Evolving activation functions for side-channel analysis.
\newblock {\em IACR Cryptol. ePrint Arch.}, 2021:249, 2021.

\bibitem{Kocher_SCA}
P.~C. Kocher, J.~Jaffe, and B.~Jun.
\newblock Differential power analysis.
\newblock In {\em Proceedings of the 19th Annual International Cryptology
  Conference on Advances in Cryptology}, CRYPTO '99, pages 388--397, London,
  UK, UK, 1999. Springer-Verlag.

\bibitem{breaking-crypto-with-dl}
H.~Maghrebi, T.~Portigliatti, and E.~Prouff.
\newblock Breaking cryptographic implementations using deep learning
  techniques.
\newblock In {\em International Conference on Security, Privacy, and Applied
  Cryptography Engineering}, pages 3--26. Springer, 2016.

\bibitem{dpa_book}
S.~Mangard, E.~Oswald, and T.~Popp.
\newblock {\em {Power Analysis Attacks: Revealing the Secrets of Smart Cards}}.
\newblock {Springer}, December 2006.

\bibitem{Miller95}
B.~L. Miller and D.~E. Goldberg.
\newblock Genetic algorithms, selection schemes, and the varying effects of
  noise.
\newblock {\em Evol. Comput.}, 4(2):113–131, jun 1996.

\bibitem{Picek_Heuser_Jovic_Bhasin_Regazzoni_2018}
S.~Picek, A.~Heuser, A.~Jovic, S.~Bhasin, and F.~Regazzoni.
\newblock The curse of class imbalance and conflicting metrics with machine
  learning for side-channel evaluations.
\newblock {\em IACR Transactions on Cryptographic Hardware and Embedded
  Systems}, 2019(1):209--237, Nov. 2018.

\bibitem{picek-sok}
S.~Picek, G.~Perin, L.~Mariot, L.~Wu, and L.~Batina.
\newblock Sok: Deep learning-based physical side-channel analysis.
\newblock {\em ACM Comput. Surv.}, oct 2022.
\newblock Just Accepted.

\bibitem{10.1007/3-540-45418-7_17}
J.-J. Quisquater and D.~Samyde.
\newblock Electromagnetic analysis (ema): Measures and counter-measures for
  smart cards.
\newblock In I.~Attali and T.~Jensen, editors, {\em Smart Card Programming and
  Security}, pages 200--210, Berlin, Heidelberg, 2001. Springer Berlin
  Heidelberg.

\bibitem{large-scale-evo-of-img-clf-2017}
E.~Real, S.~Moore, A.~Selle, S.~Saxena, Y.~L. Suematsu, J.~Tan, Q.~Le, and
  A.~Kurakin.
\newblock Large-scale evolution of image classifiers.
\newblock {\em arXiv preprint arXiv:1703.01041}, 2017.

\bibitem{Rijsdijk_Wu_Perin_Picek_2021}
J.~Rijsdijk, L.~Wu, G.~Perin, and S.~Picek.
\newblock Reinforcement learning for hyperparameter tuning in deep
  learning-based side-channel analysis.
\newblock {\em IACR Transactions on Cryptographic Hardware and Embedded
  Systems}, 2021(3):677–707, Jul. 2021.

\bibitem{vgg-16-competition-paper}
K.~Simonyan and A.~Zisserman.
\newblock Very deep convolutional networks for large-scale image recognition.
\newblock {\em arXiv preprint arXiv:1409.1556}, 2014.

\bibitem{10.1007/978-3-642-01001-9_26}
F.-X. Standaert, T.~G. Malkin, and M.~Yung.
\newblock A unified framework for the analysis of side-channel key recovery
  attacks.
\newblock In A.~Joux, editor, {\em Advances in Cryptology - EUROCRYPT 2009},
  pages 443--461, Berlin, Heidelberg, 2009. Springer Berlin Heidelberg.

\bibitem{neat-paper}
K.~O. Stanley and R.~Miikkulainen.
\newblock Evolving neural networks through augmenting topologies.
\newblock {\em Evolutionary computation}, 10(2):99--127, 2002.

\bibitem{cnn-evo-for-clf-2019}
Y.~Sun, B.~Xue, M.~Zhang, and G.~G. Yen.
\newblock Evolving deep convolutional neural networks for image classification.
\newblock {\em IEEE Transactions on Evolutionary Computation}, 24(2):394--407,
  2019.

\bibitem{dpav2-dataset}
{TELECOM ParisTech SEN research group}.
\newblock Dpa contest ($2^{nd}$ edition), 2009-2010.

\bibitem{revisiting-efficient-cnn-archs}
L.~Wouters, V.~Arribas, B.~Gierlichs, and B.~Preneel.
\newblock Revisiting a methodology for efficient cnn architectures in profiling
  attacks.
\newblock {\em IACR Transactions on Cryptographic Hardware and Embedded
  Systems}, 2020(3):147–168, Jun. 2020.

\bibitem{lichao-auto-hyperparam-tuning}
L.~Wu, G.~Perin, and S.~Picek.
\newblock I choose you: Automated hyperparameter tuning for deep learning-based
  side-channel analysis.
\newblock {\em IACR Cryptol. ePrint Arch.}, 2020:1293, 2020.

\bibitem{ae-for-sca-traces}
L.~Wu and S.~Picek.
\newblock Remove some noise: On pre-processing of side-channel measurements
  with autoencoders.
\newblock {\em IACR Transactions on Cryptographic Hardware and Embedded
  Systems}, pages 389--415, 2020.

\bibitem{cryptoeprint:2021/717}
L.~Wu, Y.-S. Won, D.~Jap, G.~Perin, S.~Bhasin, and S.~Picek.
\newblock Explain some noise: Ablation analysis for deep learning-based
  physical side-channel analysis.
\newblock Cryptology ePrint Archive, Paper 2021/717, 2021.

\bibitem{efficient-cnn-arch-method}
G.~Zaid, L.~Bossuet, A.~Habrard, and A.~Venelli.
\newblock Methodology for efficient cnn architectures in profiling attacks.
\newblock {\em IACR Transactions on Cryptographic Hardware and Embedded
  Systems}, pages 1--36, 2020.

\end{thebibliography}

\end{document}